\documentclass[runningheads]{llncs}
\usepackage[T1]{fontenc}
\usepackage{graphicx,verbatim}
\usepackage{enumitem} 
\newlist{noitemize}{itemize}{1}
\setlist[noitemize]{
    label={}, 
    labelsep=0pt, 
    leftmargin=0pt,
    before=\begin{flushleft},
    after=\end{flushleft}\vspace{\baselineskip} 
}
\usepackage{float}
\usepackage[table]{xcolor}
\usepackage{tabularx,booktabs,makecell,multirow}
\newcolumntype{Y}{>{\centering\arraybackslash}X}
\usepackage{hyperref}
\usepackage{color}

\begin{document}
\title{Agentic Large Language Models for Training-Free Neuro-Radiological Image Analysis}
\titlerunning{Agentic LLMs for Clinical Neuro-Radiology}
%



\author{%
Ayhan Can Erdur\inst{1,2,} \and
Daniel Scholz\inst{2,3,4} \and
Jiazhen Pan\inst{2,4} \and
Benedikt Wiestler\inst{3,4, \ast} \and
Daniel Rueckert\inst{2,4,5,\ast} \and 
Jan C. Peeken\inst{1,6,\ast}
}

\authorrunning{Erdur et al.}

\institute{%
Department of Radiation Oncology, TUM University Hospital, Munich, Germany \and
Chair for AI in Healthcare and Medicine, Technical University of Munich (TUM) and TUM University Hospital, Munich, Germany \and
Chair for AI for Image-Guided Diagnosis and Therapy, Technical University of Munich (TUM) and TUM University Hospital, Munich, Germany \and
Munich Center for Machine Learning (MCML), Munich, Germany \and
Department of Computing, Imperial College London, London, UK \and
Deutsches Konsortium für Translationale Krebsforschung (DKTK), Partner Site Munich, Munich, Germany \\
\textsuperscript{$\ast$} contributed equally as last authors  \\
 \email{can.erdur@.tum.de}
}  

\maketitle              
\begin{abstract}

State-of-the-art large language models (LLMs) show high performance in general visual question answering. However, a fundamental limitation remains: current architectures lack the native 3D spatial reasoning required to directly analyze volumetric medical imaging, such as CT or MRI. Emerging agentic AI offers a new solution, eliminating the need for intrinsic 3D processing by enabling LLMs to orchestrate and leverage specialized external tools. Yet, the feasibility of such agentic frameworks in complex, multi-step radiological workflows remains underexplored.
In this work, we present a training-free agentic pipeline for automated brain MRI analysis. Validating our methodology on several LLMs (GPT-5.4, Gemini 3.1 Pro, Claude Sonnet 4.6) with off-the-shelf domain-specific tools, our system autonomously executes complex end-to-end workflows, including preprocessing (skull stripping, registration), pathology segmentation (glioma, meningioma, metastases), and volumetric reasoning.
We evaluate our framework across increasingly complex radiological tasks, from single-scan tumor and anatomy analysis to longitudinal response assessment requiring multi-timepoint comparisons. We analyze the impact of architectural design by comparing single-agent models against multi-agent "domain-expert" collaborations. To support rigorous evaluation of future agentic systems, we release a benchmark dataset of image-prompt-answer tuples derived from public data.
Our results demonstrate that agentic AI can solve highly neuro-radiological image analysis tasks through tool use without the need for training or fine-tuning. 


\end{abstract}

\section{Introduction}

Imaging is central to neurological diagnosis, surgical planning, and treatment monitoring, yet extracting quantitative insights remains highly labor-intensive. Critical downstream tasks require precise delineation of anatomy and lesions across hundreds of slices, making annotation a tedious process that can consume hours of expert labor per scan.
While deep learning has enabled rapid delineation through automated segmentation \cite{erdur2025deep}, deploying and orchestrating AI models in practice often requires familiarity with coding and infrastructure setup. Furthermore, real-world clinical imaging interpretation rarely ends with a single segmentation output, but demands complex, multi-step workflows, e.g., for assessing therapy response  using Response Assessment in Neuro-Oncology (RANO) criteria. Since the burden of orchestrating these pipelines and managing intermediate artifacts falls on the user, the broader clinical adoption of these tools remains limited \cite{kelly2019key}. Attempting to automate this orchestration via rule-based scripts quickly becomes infeasible to extend, as expanding task domains, input conditions, and user needs inevitably trigger an unmaintainable amount of conditional clauses.

Recently, large language models (LLMs) have demonstrated broad capabilities across diverse tasks, including medicine \cite{thirunavukarasu2023large}, and have transformed computational workflows by offering a natural-language interface that lowers the barrier to use. This paradigm has been extended to vision \cite{eriksen2024use}, and emerging medical variants have achieved robust performance on 2D visual question-answering (VQA) tasks and diagnostic summarization \cite{li2023llava,moor2023med}. 

Despite these successes, current LLMs lack native reasoning capabilities for 3D volumetric imaging, leaving critical imaging workflows unsupported. The AI field is shifting toward agentic systems that leverage external algorithms as functional tools \cite{schick2023toolformer}. This autonomous orchestration offers a powerful mechanism to bridge the 3D gap by integrating specialized medical image processing into coherent clinical pipelines. Yet, application of such agentic workflows to neuro-radiology remains largely unexplored.

To bridge the gap between LLMs and imaging analysis in clinical practice, we make the following contributions:
\begin{itemize}[leftmargin=10pt, labelsep=5pt, label=$\bullet$]
    \item We introduce an end-to-end brain MRI analysis pipeline that orchestrates preprocessing, segmentation, volumetry, and tool-grounded clinical conclusions without any training/fine-tuning.
    \item We design and benchmark multiple agentic architectures, comparing single-
    agent and multi-agent variants for single- and multi-timepoint volumetric analysis and RANO assessment. 
    \item Our results demonstrate the robustness of our framework across multiple all-purpose LLMs (GPT-5.4, Gemini 3.1 Pro, Claude Sonnet 4.6), proving that high performance does not require specialized medical experts.
    \item We release a brain MRI VQA benchmark dataset to enable reproducible evaluation and future extensions.\\ \url{https://github.com/RadOnc-AI-group/brain-mri-agents}
\end{itemize}

\section{Related Work}

\noindent \textbf{Agentic AI in Medical Imaging} Recent LLMs can process 3D medical volumes \cite{bai2024m3d,xin2025med3dvlm}, but native 3D modeling is costly and often imprecise in spatial localization \cite{liu2025comparison}. As a result, the field is moving toward agentic AI, where language models utilize external tools. MedRAX \cite{fallahpour2025medrax} applies this to 2D chest X-ray VQA, TissueLab \cite{li2025co} extends tool-based quantification across pathology and radiology, and MedAgentSim \cite{almansoori2025medagentsim} models multi-agent diagnostic dialogue. These systems demonstrate orchestration in 2D imaging, generalized quantification, and dialogue, whereas we focus on training-free LLM orchestration for operational neuro-radiology in multi-step 3D brain MRI workflows. \\

\noindent \textbf{Benchmarking in Neuro-Oncology Workflows} The application of agentic AI necessitates benchmarks that evaluate task orchestration.
Recent benchmarks evaluating language models in this domain, such as OmniBrainBench \cite{peng2025omnibrainbench} and BraTS-derived QA \cite{safari2025performance}, measure static, single-step VQA on 2D representations. While they test visual comprehension, they do not assess an agent's ability to plan or to iteratively invoke tools, nor do they focus on 3D.


\section{Materials and Methods}
\begin{figure}
\centering
\includegraphics[width=
\linewidth]{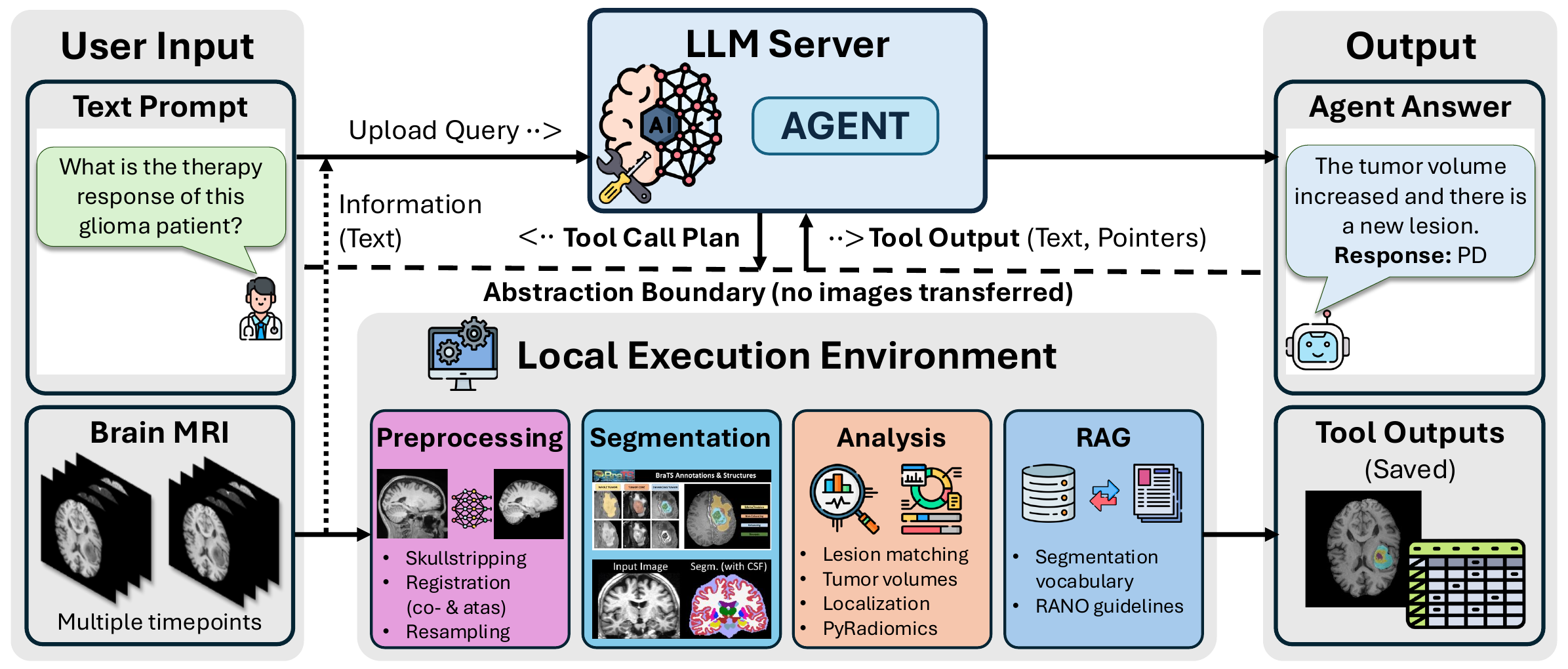}
\caption{Overview of the agentic brain MRI analysis pipeline. Only text objects are uploaded to the LLM server; images are kept local. Outputs from segmentation operations (lesion matching, volume extraction) are provided to the agent in an informative text.} \label{fig:workflow}
\end{figure}

We present an agentic framework that operates on 3D MRI volumes in response to natural language queries. Given a clinical prompt and the corresponding imaging data, the LLM agent plans and autonomously executes the required tools, returning intermediate outputs (e.g., segmentation and  measurement files) along with the text response for continued interaction (Fig.~\ref{fig:workflow}).
All tools run locally such that the LLM can interact only via text-based inputs/outputs and object pointers, ensuring sensitive imaging data is never transmitted to commercial servers. The framework is built using the OpenAI Agents SDK. The \texttt{temperature} parameter for the LLM is set to 0.

\subsection{Available Tools}
Our toolset consists  of 24 tools for the agent's usage, spanning from specialized neural networks to Python libraries and custom helper functions: \\
\textbf{Preprocessing}: SynthStrip \cite{hoopes2022synthstrip} skull stripping; ANTsPy \cite{avants2011ants} registration (T1 co-registration; atlas registration to SRI24/MNI152); resampling.\\
\textbf{Segmentation}: BraTS winner models (glioma, metastases, meningioma, pediatric) via BraTS Orchestrator \cite{kofler2025bratsorchestrator}; SynthSeg \cite{billot2023synthseg} for 32 anatomical regions. \\
\textbf{Analysis}: lesion enumeration (connected components); IoU-based longitudinal lesion matching (Panoptica \cite{kofler2023panoptica});  lesion-wise sub-volume and geometry (bounding boxes/ centroids) extraction; morphological and textural feature extraction (PyRadiomics \cite{van2017pyradiomics}); lesion localization (USCLobes atlas \cite{joshi2022hybrid}). \\
\textbf{Utilities}: registration verification by comparing image headers/metadata across scans and atlas templates; image loading; Retrieval Augmented Generation (RAG) for brain-region classes and labels, and RANO criteria (to reduce default input tokens by on-demand access); and visualization.\\

There are two fundamental 'philosophies' in tool design: bundling multiple operations into a single tool call to maximize reliability and reduce orchestration overhead, or exposing individual, atomic tools (e.g.,\texttt{load-image}) to let the agent decide what to run and when. We follow the latter, as it avoids redundant actions and unnecessary intermediate products, provides a more faithful stress test for agentic planning, and enables easier future integration of new tools.

A board-certified radiologist adapted the RANO criteria \cite{wen2023rano} for RAG by distilling them into a Markdown document focused on image-based rules, omitting corticosteroid dosages and clinical status. 

\subsection{Agent Architectures} We investigate three variants of agent architectures. In the \emph{single-agent} setting, one agent is responsible for both workflow planning and execution. This requires providing the agent with access to all tools, exposing every tool description to the LLM as input tokens. To reduce the burden on a monolithic agent and inject only task-relevant knowledge at each call, we also consider multi-agent setups with specialized experts for preprocessing, segmentation, and analysis. Coordination between these agents is implemented in two ways: (i) \emph{agents-as-tools} and (ii) \emph{handoffs}. 

\begin{figure}
\centering
\includegraphics[width=\textwidth]{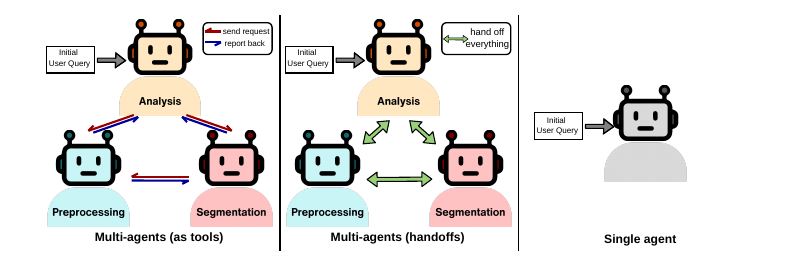}
\caption{Different designs of multi-agent setup} \label{fig:agents}
\end{figure}

In \emph{agents-as-tools}, agents are introduced as callable tools and explicitly dictate tasks to one another, introducing a slight hierarchy. Communication in this setup relies on well-formatted \texttt{Request} and \texttt{Response} schemes rather than free-text messaging. \emph{handoffs} follows an implicit approach in which control is transferred by passing the full execution context and history to another agent, granting more autonomy to each agent. 
\autoref{fig:agents} visualizes the agent dynamics. 

While tools are naturally assigned based on their primary task space (e.g., segmentation tools to the segmentation agent), other allocations are more nuanced. Specifically, we assign \texttt{registration-verification} to the segmentation agent to enable the local evaluation of network-specific prerequisites. Placing this tool elsewhere would require sharing prerequisite knowledge across agents and increase costs due to extra inter-agent routing. The remaining utility tools are assigned to the analysis agent.

\subsection{Dataset} We construct a brain MRI VQA dataset to benchmark our agent architectures and LLMs. Each entry comprises a free-text question, an expected tool-call plan, and an expected keyword-value answer. For each of the three architectures, we define a corresponding \emph{ground-truth} tool sequence that includes agent \emph{handoffs} and requests.

We  generate multiple QA pairs per patient and partition the dataset into two subtasks of increasing workflow complexity. Queries range from single-field outputs to detailed per-lesion analyses that require reporting up to 40 fields. \\\\
\textbf{(Task 1) Single-timepoint assessment}: post-segmentation measurements (e.g., volumes, locations); 50 patients / 150 cases (51 preprocessing) / 769 queries.
Avg. \#actions: \emph{single} 5.10; \emph{agents-as-tools}/\emph{handoffs} 7.63 \\
\textbf{(Task 2.1) Longitudinal response analysis}: multi-timepoint extension of single-timepoint analysis (e.g., volume change, new lesions); 25 patients / 75 cases (21 preprocessing) / 809 queries.\\
Avg. \#actions: \emph{single} 11.67; \emph{agents-as-tools}/\emph{handoffs} 13.11 \\
\textbf{(Task 2.2) RANO assessment}: 30 patients / 30 cases / 30 queries. \\


Our imaging data is a multi-centric cohort comprising patients with pre-operative glioma (n=20) \cite{baid2021bratsgli,van2021erasmus,bakas2022upenngbm,calabrese2022ucsfgbm}, post-operative glioma (n=42) \cite{de20242bratspost,suter2022lumiere}, metastasis (n=24) \cite{moawad2024bratsmets,rudie2024ucsfmets,flouri2025proteas}, and meningioma (n=7) \cite{labella2023bratsmen}. All include T1, T2, T1ce, and FLAIR scans. A subset (10, 35, 10, and 0, respectively) includes 2--3 timepoints for longitudinal analysis, comprising 30 patients from the LUMIERE \cite{suter2022lumiere} dataset with expert RANO ratings. We randomly sample 2 post-operative image sets to test the follow-up RANO assessment.

While most images are skull-stripped and preprocessed, 11 require registration to the SRI24 atlas. To test our preprocessing pipeline at full capacity, we add unprocessed in-house pre-operative glioma (n=8) and metastasis (n=4) cases from TUM University Hospital (research waiver obtained). 


\subsection{Evaluation} Evaluating agentic workflows remains an open challenge \cite{yehudai2025survey}. We judge performance along three dimensions: \emph{tool-call fidelity} (precision, recall, number of erroneous tool calls), \emph{computational cost} (input/output tokens), and \emph{response quality} (query inclusion rate, accuracy).

Precision and recall for tool-call fidelity are measured as order-invariant metrics because valid workflows can differ in step order (e.g., retrieving the RANO criteria before or after segmentation is negligible). Number of errors captures missing/wrong preprocessing and segmentation steps that can affect the final response.

We evaluate the agents' unstructured free-text responses using a hybrid approach that combines an LLM judge (OpenAI GPT-5.5) with deterministic verification and human oversight. First, the LLM judge parses the free-text output into discrete key-value pairs based on ground-truth keys and computes an inclusion rate by counting the successfully extracted keys. The accuracy of the corresponding values is then verified via deterministic scripts. To account for the ambiguity of LLM-based evaluation, a human expert conducts a manual review of a fixed 30\% of cases per run, assessing both query inclusion and value correctness.



\section{Results and Discussion}

\begin{table}[!ht]
\caption{Performance of different agent architectures and LLMs on single-timepoint volumetry (Task 1, 150 cases, 769 queries). Green cells highlight the best overall result in the task, and the bold text shows the best local performance for each LLM.}
\label{tab:results-task2}
{\fontsize{8pt}{10pt}\selectfont
\begin{tabularx}{\textwidth}{llccYYYYYYYY}
\toprule
\textbf{LLM} & \makecell{\textbf{Agent}\\ \textbf{Type}}  & \multicolumn{4}{c}{\textbf{Tool Call Fidelity}} & \multicolumn{2}{c}{\makecell{\textbf{Average}\\ \textbf{Run Costs}}} & \multicolumn{2}{c}{\makecell{\textbf{Response}\\ \textbf{Quality}\\(LLM Judge)}} & \multicolumn{2}{c}{\makecell{\textbf{Response}\\ \textbf{Quality}\\(Human)}} \\
\cline{1-12}
 &  & \multicolumn{2}{c}{\#Errors} & \multirow[c]{2}{*}{Prec.} & \multirow[c]{2}{*}{Rec.} & \multicolumn{2}{c}{Tokens} & Incl. & \multirow[c]{2}{*}{Acc.} & Incl. & \multirow[c]{2}{*}{Acc.} \\
 &  & Prep. & Seg. &  &  & Input & Out & Rate &  & Rate &  \\
\midrule
\multirow[c]{3}{*}{\parbox[c]{1.2cm}{Claude\\Sonnet 4.6}} & as-tools & 17 & 2 & 0.923 & 0.989 & \textbf{30.0K} & 1.4K & 0.993 & 0.932 & \textbf{\cellcolor{green!30}1.0} & 0.891 \\
 & handoffs & \textbf{\cellcolor{green!30}0} & 1 & \textbf{0.960} & \textbf{0.998} & 33.2K & 1.2K & \textbf{\cellcolor{green!30}1.0} & \textbf{\cellcolor{green!30}0.984} & 0.992 & \textbf{0.952} \\
 & single & \textbf{\cellcolor{green!30}0} & \textbf{\cellcolor{green!30}0} & 0.953 & 0.996 & 36.6K & \textbf{975} & \textbf{\cellcolor{green!30}1.0} & 0.977 & 0.967 & 0.943 \\
\midrule
\multirow[c]{3}{*}{\parbox[c]{1.2cm}{GPT\\5.4}} & as-tools & \textbf{\cellcolor{green!30}0} & \textbf{\cellcolor{green!30}0} & 0.990 & \textbf{\cellcolor{green!30}0.999} & 17.7K & 597 & 0.995 & 0.922 & 0.978 & 0.875 \\
 & handoffs & 1 & \textbf{\cellcolor{green!30}0} & \textbf{\cellcolor{green!30}0.993} & 0.993 & \textbf{\cellcolor{green!30}16.6K} & 435 & 0.991 & \textbf{0.956} & \textbf{\cellcolor{green!30}1.0} & \textbf{\cellcolor{green!30}0.958} \\
 & single & 2 & \textbf{\cellcolor{green!30}0} & 0.983 & \textbf{\cellcolor{green!30}0.999} & 21.5K & \textbf{\cellcolor{green!30}391} & \textbf{0.997} & 0.943 & \textbf{\cellcolor{green!30}1.0} & \textbf{\cellcolor{green!30}0.958} \\
\midrule
\multirow[c]{3}{*}{\parbox[c]{1.2cm}{Gemini\\3.1 Pro}} & as-tools & 26 & 6 & 0.829 & 0.982 & 31.5K & 1.1K & 0.993 & 0.858 & \textbf{0.975} & 0.819 \\
 & handoffs & 1 & \textbf{\cellcolor{green!30}0} & 0.963 & \textbf{0.998} & \textbf{30.1K} & 653 & 0.999 & \textbf{0.965} & 0.967 & 0.919 \\
 & single & \textbf{\cellcolor{green!30}0} & 1 & \textbf{0.981} & \textbf{0.998} & 32.3K & \textbf{540} & \textbf{\cellcolor{green!30}1.0} & 0.956 & 0.967 & \textbf{0.920} \\

\bottomrule

\end{tabularx}
}
\end{table}

Across all tasks, intent recognition remains high, with query inclusion rates approaching 1.0 (Tables~\ref{tab:results-task2} and \ref{tab:results-task3}). This shows that the models consistently understand user requests, plan workflows accurately, and deliver comprehensive reports. Human evaluation aligns with the LLM judge's scores, further validating our methods. Performance drops slightly in Task 2, where longitudinal dependencies make outputs vulnerable to error propagation.

The \emph{single} framework demonstrates robust performance by handling the extensive context across the entire toolset. However, it is the most expensive architecture because it must ingest the full tool description for every API call. \emph{handoffs} emerges as the practical compromise: it matches or exceeds the \emph{single} baseline's accuracy while maintaining lower input and output token costs. 

\emph{as-tools} architecture underperforms across nearly all models, revealing a bottleneck in explicit agent communication. Agents apply unnecessary preprocessing steps (such as redundant skull-stripping), thereby altering the images and degrading downstream segmentation accuracy. Gemini 3.1 Pro, with this architecture, also suffers from incorrect segmentation tool choice, particularly for post-operative glioma cases, resulting in 21 segmentation errors in Task 2. The large drop in human-judged accuracy is attributed to the prominence of these faulty cases in the evaluated subset. 

\begin{table}[!ht]
\caption{Performance of different agent architectures and LLMs on longitudinal response analysis without RANO (Task 2, 75 cases, 809 queries). Green cells highlight the best overall result in the task, and the bold text shows the best local performance for each LLM.}
\label{tab:results-task3}
{\fontsize{8pt}{10pt}\selectfont
\begin{tabularx}{\textwidth}{llccYYYYYYYY}
\toprule
\textbf{LLM} & \makecell{\textbf{Agent}\\ \textbf{Type}}  & \multicolumn{4}{c}{\textbf{Tool Call Fidelity}} & \multicolumn{2}{c}{\makecell{\textbf{Average}\\ \textbf{Run Costs}}} & \multicolumn{2}{c}{\makecell{\textbf{Response}\\ \textbf{Quality}\\(LLM Judge)}} & \multicolumn{2}{c}{\makecell{\textbf{Response}\\ \textbf{Quality}\\(Human)}} \\
\cline{1-12}
 &  & \multicolumn{2}{c}{\#Errors} & \multirow[c]{2}{*}{Prec.} & \multirow[c]{2}{*}{Rec.} & \multicolumn{2}{c}{Tokens} & Incl. & \multirow[c]{2}{*}{Acc.} & Incl. & \multirow[c]{2}{*}{Acc.} \\
 &  & Prep. & Seg. &  &  & Input & Out & Rate &  & Rate &  \\
\midrule
\multirow[c]{3}{*}{\parbox[c]{1.2cm}{Claude\\Sonnet 4.6}} & as-tools & 3 & 14 & 0.837 & 0.980 & 58.0K & 3.5K & \textbf{\cellcolor{green!30}1.0} & 0.859 & 0.943 & 0.776 \\
 & handoffs & \textbf{\cellcolor{green!30}0} & \textbf{\cellcolor{green!30}0} & 0.966 & 0.987 & \textbf{49.7K} & 2.5K & 0.986 & \textbf{0.907} & \textbf{0.961} & \textbf{0.891} \\
 & single & 2 & \textbf{\cellcolor{green!30}0} & \textbf{0.990} & \textbf{\cellcolor{green!30}0.998} & 56.9K & \textbf{2.3K} & 0.990 & 0.886 & 0.955 & 0.878 \\
\midrule
\multirow[c]{3}{*}{\parbox[c]{1.2cm}{GPT\\5.4}} & as-tools & 3 & 5 & 0.993 & \textbf{0.985} & 27.0K & 1.6K & \textbf{\cellcolor{green!30}1.0} & 0.889 & 0.963 & 0.896 \\
 & handoffs & \textbf{\cellcolor{green!30}0} & \textbf{\cellcolor{green!30}0} & 0.993 & \textbf{0.985} & \textbf{\cellcolor{green!30}26.0K} & 1.2K & 0.986 & \textbf{\cellcolor{green!30}0.925} & \textbf{0.970} & \textbf{\cellcolor{green!30}0.913} \\
 & single & \textbf{\cellcolor{green!30}0} & \textbf{\cellcolor{green!30}0} & \textbf{\cellcolor{green!30}0.998} & 0.979 & 30.9K & \textbf{\cellcolor{green!30}1.1K} & 0.970 & 0.909 & 0.963 & 0.874 \\
\midrule
\multirow[c]{3}{*}{\parbox[c]{1.2cm}{Gemini\\3.1 Pro}} & as-tools & 34 & 21 & 0.814 & 0.969 & 45.9K & 2.2K & 0.998 & 0.812 & 0.935 & 0.737 \\
 & handoffs & 8 & \textbf{\cellcolor{green!30}0} & 0.953 & 0.983 & \textbf{44.7K} & 1.5K & 0.986 & 0.914 & \textbf{\cellcolor{green!30}0.977} & 0.887 \\
 & single & \textbf{2} & \textbf{\cellcolor{green!30}0} & \textbf{0.988} & \textbf{0.996} & 45.1K & \textbf{1.4K} & \textbf{0.999} & \textbf{0.924} & 0.976 & \textbf{0.908} \\
\bottomrule
\end{tabularx}
}
\end{table}

Output correctness is ultimately bounded by the underlying tools. For example, the top-performing GPT-5.4 \emph{handoffs} setup achieves perfect tool-call fidelity (1.0 precision and recall) in 64 workflows, yet 17 of them fail to reach a 1.0 final response accuracy.


For the RANO assessment (Table~\ref{tab:rano_accuracy}), GPT 5.4 and Claude Sonnet 4.6 with a \emph{handoffs} or \emph{single} architecture achieve peak accuracy (0.867). These frameworks maintain a low false-negative rate, misclassifying only 2 out of 18 \textit{Progressive Disease} (PD) cases. 
A common pitfall for all configurations occurs with a patient with zero tumor volume at two successive timepoints. The agents misinterpret the empty follow-up as a CR, whereas the LUMIERE ground truth denotes SD.

\begin{table}[!ht]
\setlength{\tabcolsep}{4pt} 
\caption{RANO assessment accuracy of agent architectures and LLMs (30 cases). All cases are human-evaluated.}
{\fontsize{8pt}{10pt}\selectfont
\begin{tabularx}{\textwidth}{l cYY cYY cYY}
\toprule
& \multicolumn{3}{c}{\textbf{Claude Sonnet 4.6}} & \multicolumn{3}{c}{\textbf{GPT 5.4}} & \multicolumn{3}{c}{\textbf{Gemini 3.1 Pro}} \\
\cmidrule(lr){2-4} \cmidrule(lr){5-7} \cmidrule(lr){8-10}
 & as-tools & handoffs & single & as-tools & handoffs & single & as-tools & handoffs & single\\
\midrule
Accuracy & 0.733 & \textbf{0.867} & 0.833 & 0.767 & \textbf{0.867} & \textbf{0.867} & 0.7 & 0.767 & 0.8 \\
\bottomrule
\end{tabularx}
}
\label{tab:rano_accuracy}
\end{table}

Key behavioral discrepancies emerge between the models. Claude Sonnet 4.6 is more expensive across all tasks due to its distinct tokenizer and inherently verbose, "chattier" output style. While Claude excels at single-timepoint volumetry, it is outperformed by other LLMs in longitudinal analysis. However, it demonstrates stricter adherence to complex RANO guidelines than Gemini. Despite these trends, no LLM emerges as the definitive best across all architectures and tasks.

\section{Conclusion}

As the adoption of LLMs in clinical settings becomes inevitable, we propose a training-free agentic framework that safely bridges the gap between 3D medical imaging and general-purpose foundation models. It offers robust workflow assistance to radiologists without exposing sensitive patient data to commercial servers, at a scalable cost of 5–20 cents per run, depending on API pricing.

Systematic benchmarking shows that our handoff-based multi-agent configuration achieves the highest overall performance and efficiency, with the single-agent baseline closely tailing it, while the agents-as-tools approach is subpar.
These architectural trends hold true across all evaluated models, underscoring the LLM-agnostic robustness.

Our framework presents a few limitations. First, overall performance and latency, as well as the scope of supported workflows, are inherently constrained by the underlying tools. Second, this study focuses only on brain MRI, but the system's modular design enables future extension into other domains and modalities. Finally, our pipeline does not yet support multi-tiered, cascaded decision systems, such as classifying pathologies before selecting disease-specific downstream tools, which remains an active area of research.

We demonstrate the capabilities of agentic AI systems in real-world clinical workflows, establish a practical blueprint for deploying autonomous agents in medical imaging, and introduce a new VQA dataset to enable future research. The complete code repository and dataset are released in the provided link.

\begin{credits}
\subsubsection{\ackname} This study was supported by the DFG, grant \#504320104.

\subsubsection{\discintname}
The authors declare no competing interests relevant to the content in this article.
\end{credits}

%
%
%
\bibliographystyle{splncs04}
\bibliography{bibliography}

@article{erdur2025deep,
  title={Deep learning for autosegmentation for radiotherapy treatment planning: State-of-the-art and novel perspectives},
  author={Erdur, Ayhan Can and Rusche, Daniel and Scholz, Daniel and Kiechle, Johannes and Fischer, Stefan and Llorian-Salvador, Oscar and others},
  journal={Strahlentherapie und Onkologie},
  volume={201},
  number={3},
  pages={236--254},
  year={2025},
  publisher={Springer}
}

@article{kelly2019key,
  title={Key challenges for delivering clinical impact with artificial intelligence},
  author={Kelly, Christopher J and Karthikesalingam, Alan and Suleyman, Mustafa and Corrado, Greg and King, Dominic},
  journal={BMC medicine},
  volume={17},
  number={1},
  pages={195},
  year={2019},
  publisher={Springer}
}

@article{thirunavukarasu2023large,
  title={Large language models in medicine},
  author={Thirunavukarasu, Arun James and Ting, Darren Shu Jeng and Elangovan, Kabilan and Gutierrez, Laura and Tan, Ting Fang and Ting, Daniel Shu Wei},
  journal={Nature medicine},
  volume={29},
  number={8},
  pages={1930--1940},
  year={2023},
  publisher={Nature Publishing Group US New York}
}

@misc{eriksen2024use,
  title={Use of GPT-4 to diagnose complex clinical cases},
  author={Eriksen, Alexander V and M{\"o}ller, S{\"o}ren and Ryg, Jesper},
  journal={Nejm Ai},
  volume={1},
  number={1},
  pages={AIp2300031},
  year={2024},
  publisher={Massachusetts Medical Society}
}

@article{li2023llava,
  title={Llava-med: Training a large language-and-vision assistant for biomedicine in one day},
  author={Li, Chunyuan and Wong, Cliff and Zhang, Sheng and Usuyama, Naoto and Liu, Haotian and Yang, Jianwei and others},
  journal={NeurIPS},
  volume={36},
  pages={28541--28564},
  year={2023}
}

@inproceedings{moor2023med,
  title={Med-flamingo: a multimodal medical few-shot learner},
  author={Moor, Michael and Huang, Qian and Wu, Shirley and Yasunaga, Michihiro and Dalmia, Yash and Leskovec, Jure and others},
  booktitle={Machine learning for health (ML4H)},
  pages={353--367},
  year={2023},
  organization={PMLR}
}

@article{schick2023toolformer,
  title={Toolformer: Language models can teach themselves to use tools},
  author={Schick, Timo and Dwivedi-Yu, Jane and Dess{\`\i}, Roberto and Raileanu, Roberta and Lomeli, Maria and Hambro, Eric and others},
  journal={NeurIPS},
  volume={36},
  pages={68539--68551},
  year={2023}
}

@article{hoopes2022synthstrip,
  title={SynthStrip: skull-stripping for any brain image},
  author={Hoopes, Andrew and Mora, Jocelyn S and Dalca, Adrian V and Fischl, Bruce and Hoffmann, Malte},
  journal={NeuroImage},
  volume={260},
  pages={119474},
  year={2022},
  publisher={Elsevier}
}

@article{avants2011ants,
  title={A reproducible evaluation of ANTs similarity metric performance in brain image registration},
  author={Avants, Brian B and Tustison, Nicholas J and Song, Gang and Cook, Philip A and Klein, Arno and Gee, James C},
  journal={NeuroImage},
  volume={54},
  number={3},
  pages={2033--2044},
  year={2011},
  publisher={Elsevier}
}

@article{baid2021bratsgli,
  title={The rsna-asnr-miccai brats 2021 benchmark on brain tumor segmentation and radiogenomic classification},
  author={Baid, Ujjwal and Ghodasara, Satyam and Mohan, Suyash and Bilello, Michel and Calabrese, Evan and Colak, Errol and others},
  journal={arXiv:2107.02314},
  year={2021}
}

@article{moawad2024bratsmets,
  title={The brain tumor segmentation-metastases (brats-mets) challenge 2023: Brain metastasis segmentation on pre-treatment mri},
  author={Moawad, Ahmed W and Janas, Anastasia and Baid, Ujjwal and Ramakrishnan, Divya and Saluja, Rachit and Ashraf, Nader and others},
  journal={arxiv},
  pages={arXiv--2306},
  year={2024}
}

@article{de20242bratspost,
  title={The 2024 brain tumor segmentation (brats) challenge: Glioma segmentation on post-treatment mri},
  author={de Verdier, Maria Correia and Saluja, Rachit and Gagnon, Louis and LaBella, Dominic and Baid, Ujjwall and Tahon, Nourel Hoda and others},
  journal={arXiv:2405.18368},
  year={2024}
}

@article{labella2023bratsmen,
  title={The asnr-miccai brain tumor segmentation (brats) challenge 2023: Intracranial meningioma},
  author={LaBella, Dominic and Adewole, Maruf and Alonso-Basanta, Michelle and Altes, Talissa and Anwar, Syed Muhammad and Baid, Ujjwal and others},
  journal={arXiv:2305.07642},
  year={2023}
}

@article{van2021erasmus,
  title={The Erasmus Glioma Database (EGD): Structural MRI scans, WHO 2016 subtypes, and segmentations of 774 patients with glioma},
  author={van der Voort, Sebastian R and Incekara, Fatih and Wijnenga, Maarten MJ and Kapsas, Georgios and Gahrmann, Renske and Schouten, Joost W and others},
  journal={Data in brief},
  volume={37},
  pages={107191},
  year={2021},
  publisher={Elsevier}
}

@article{bakas2022upenngbm,
  title={The University of Pennsylvania glioblastoma (UPenn-GBM) cohort: advanced MRI, clinical, genomics, \& radiomics},
  author={Bakas, Spyridon and Sako, Chiharu and Akbari, Hamed and Bilello, Michel and Sotiras, Aristeidis and Shukla, Gaurav and others},
  journal={Scientific data},
  volume={9},
  number={1},
  pages={453},
  year={2022},
  publisher={Nature Publishing Group UK London}
}

@article{calabrese2022ucsfgbm,
  title={The University of California San Francisco preoperative diffuse glioma MRI dataset},
  author={Calabrese, Evan and Villanueva-Meyer, Javier E and Rudie, Jeffrey D and Rauschecker, Andreas M and Baid, Ujjwal and Bakas, Spyridon and others},
  journal={Radiology: Artificial Intelligence},
  volume={4},
  number={6},
  pages={e220058},
  year={2022},
  publisher={Radiological Society of North America}
}

@article{suter2022lumiere,
  title={The LUMIERE dataset: Longitudinal Glioblastoma MRI with expert RANO evaluation},
  author={Suter, Yannick and Knecht, Urspeter and Valenzuela, Waldo and Notter, Michelle and Hewer, Ekkehard and Schucht, Philippe and others},
  journal={Scientific data},
  volume={9},
  number={1},
  pages={768},
  year={2022},
  publisher={Nature Publishing Group UK London}
}

@article{rudie2024ucsfmets,
  title={The university of California San Francisco brain metastases stereotactic radiosurgery (UCSF-BMSR) MRI dataset},
  author={Rudie, Jeffrey D and Saluja, Rachit and Weiss, David A and Nedelec, Pierre and Calabrese, Evan and Colby, John B and others},
  journal={Radiology: Artificial Intelligence},
  volume={6},
  number={2},
  pages={e230126},
  year={2024},
  publisher={Radiological Society of North America}
}

@article{flouri2025proteas,
  title={A longitudinal MRI dataset of brain metastases with tumor segmentations, clinical \& radiomic data},
  author={Flouri, Dimitra and Papanikas, Christos-Panagiotis and Manikis, Georgios C and Ioannou, Eleftherios and Karaoli, Georgia and Papageorgiou, Elisavet and others},
  journal={Scientific Data},
  volume={12},
  number={1},
  pages={1828},
  year={2025},
  publisher={Nature Publishing Group UK London}
}

@misc{kofler2025bratsorchestrator,
      title={BraTS orchestrator : Democratizing and Disseminating state-of-the-art brain tumor image analysis}, 
      author={Florian Kofler and Marcel Rosier and Mehdi Astaraki and Ujjwal Baid and Hendrik Möller and others},
      year={2025},
      eprint={2506.13807},
      archivePrefix={arXiv},
      primaryClass={eess.IV},
      url={https://arxiv.org/abs/2506.13807}, 
}

@article{billot2023synthseg,
  title={SynthSeg: Segmentation of brain MRI scans of any contrast and resolution without retraining},
  author={Billot, Benjamin and Greve, Douglas N and Puonti, Oula and Thielscher, Axel and Van Leemput, Koen and Fischl, Bruce and others},
  journal={Medical image analysis},
  volume={86},
  pages={102789},
  year={2023},
  publisher={Elsevier}
}

@article{van2017pyradiomics,
  title={Computational radiomics system to decode the radiographic phenotype},
  author={Van Griethuysen, Joost JM and Fedorov, Andriy and Parmar, Chintan and Hosny, Ahmed and Aucoin, Nicole and Narayan, Vivek and others},
  journal={Cancer research},
  volume={77},
  number={21},
  pages={e104--e107},
  year={2017},
  publisher={American Association for Cancer Research}
}

@misc{kofler2023panoptica,
      title={Panoptica -- instance-wise evaluation of 3D semantic and instance segmentation maps}, 
      author={Florian Kofler and Hendrik Möller and Josef A. Buchner and Ezequiel de la Rosa and Ivan Ezhov and Marcel Rosier and others},
      year={2023},
      eprint={2312.02608},
      archivePrefix={arXiv},
      primaryClass={cs.CV}
}

@article{fallahpour2025medrax,
  title={Medrax: Medical reasoning agent for chest x-ray},
  author={Fallahpour, Adibvafa and Ma, Jun and Munim, Alif and Lyu, Hongwei and Wang, Bo},
  journal={arXiv:2502.02673},
  year={2025}
}

@article{xin2025med3dvlm,
  title={Med3dvlm: An efficient vision-language model for 3d medical image analysis},
  author={Xin, Yu and Ates, Gorkem Can and Gong, Kuang and Shao, Wei},
  journal={IEEE Journal of Biomedical and Health Informatics},
  year={2025},
  publisher={IEEE}
}

@article{bai2024m3d,
  title={M3d: Advancing 3d medical image analysis with multi-modal large language models},
  author={Bai, Fan and Du, Yuxin and Huang, Tiejun and Meng, Max Q-H and Zhao, Bo},
  journal={arXiv:2404.00578},
  year={2024}
}

@article{li2025co,
  title={A co-evolving agentic ai system for medical imaging analysis},
  author={Li, Songhao and Xu, Jonathan and Bao, Tiancheng and Liu, Yuxuan and Liu, Yuchen and Liu, Yihang and others},
  journal={arXiv:2509.20279},
  year={2025}
}

@inproceedings{almansoori2025medagentsim,
  title={MedAgentSim: Self-evolving Multi-agent Simulations for Realistic Clinical Interactions},
  author={Almansoori, Mohammad and Kumar, Komal and Cholakkal, Hisham},
  booktitle={MICCAI},
  pages={362--372},
  year={2025},
  organization={Springer}
}

@article{safari2025performance,
  title={Performance of GPT-5 in Brain Tumor MRI Reasoning},
  author={Safari, Mojtaba and Wang, Shansong and Hu, Mingzhe and Eidex, Zach and Li, Qiang and Yang, Xiaofeng},
  journal={arXiv:2508.10865},
  year={2025}
}

@article{liu2025comparison,
  title={A Comparison and Evaluation of Fine-tuned Convolutional Neural Networks to Large Language Models for Image Classification and Segmentation of Brain Tumors on MRI},
  author={Liu, Felicia and Yoo, Jay J and Khalvati, Farzad},
  journal={arXiv:2509.10683},
  year={2025}
}

@article{peng2025omnibrainbench,
  title={OmniBrainBench: A Comprehensive Multimodal Benchmark for Brain Imaging Analysis Across Multi-stage Clinical Tasks},
  author={Peng, Zhihao and Wang, Cheng and Liu, Shengyuan and Liang, Zhiying and Ye, Zanting and Ju, Minjie and others},
  journal={arXiv:2511.00846},
  year={2025}
}

@article{yehudai2025survey,
  title={Survey on evaluation of llm-based agents},
  author={Yehudai, Asaf and Eden, Lilach and Li, Alan and Uziel, Guy and Zhao, Yilun and Bar-Haim, Roy and others},
  journal={arXiv:2503.16416},
  year={2025}
}

@article{joshi2022hybrid,
  title={A hybrid high-resolution anatomical MRI atlas with sub-parcellation of cortical gyri using resting fMRI},
  author={Joshi, Anand A and Choi, Soyoung and Liu, Yijun and Chong, Minqi and Sonkar, Gaurav and Gonzalez-Martinez, Jorge and  others},
  journal={Journal of neuroscience methods},
  volume={374},
  pages={109566},
  year={2022},
  publisher={Elsevier}
}

@article{wen2023rano,
  title={RANO 2.0: update to the response assessment in neuro-oncology criteria for high-and low-grade gliomas in adults},
  author={Wen, Patrick Y and Van Den Bent, Martin and Youssef, Gilbert and Cloughesy, Timothy F and Ellingson, Benjamin M and Weller, Michael others},
  journal={Journal of Clinical Oncology},
  volume={41},
  number={33},
  pages={5187--5199},
  year={2023},
  publisher={Wolters Kluwer Health}
}

\end{document}